\documentclass[accepted]{uai2025} 

\usepackage[american]{babel}

\usepackage{natbib} 
\usepackage{mathtools} 
\usepackage{booktabs} 
\usepackage{tikz} 
\usepackage{amsmath, amsfonts, amssymb,amsthm,multicol,derivative,enumitem,graphicx,float,tikz,tikz-qtree,forest,verbatim,bm,booktabs,lipsum}
\usepackage{algpseudocode}
\usepackage[ruled,linesnumbered]{algorithm2e}
\usetikzlibrary{fit,bayesnet, positioning, quotes,calc}
\usepackage[makeroom]{cancel}
\theoremstyle{definition}

\theoremstyle{definition}

\theoremstyle{definition}

\theoremstyle{definition}
\newtheorem{rmrk}{Remark}
\theoremstyle{definition}

\theoremstyle{definition}

\theoremstyle{definition}

\theoremstyle{definition}

\definecolor{lightblue}{RGB}{173,216,230} 
\definecolor{lightgray}{gray}{.97}

\renewcommand{\d}{\mathrm{d}}
\newcommand{\MLP}{\mathrm{MLP}}
\newcommand{\ReLU}{\mathrm{ReLU}}

\let\oldforall\forall
\renewcommand{\forall}{\oldforall \, }
\let\oldLeftrightarrow\Leftrightarrow
\renewcommand{\Leftrightarrow}{\oldLeftrightarrow \,\, }
\let\oldexist\exists
\renewcommand{\exists}{\oldexist \: }

\renewcommand{\and}{\mathrm{\,\,and\,\,}}

\newcommand{\KL}{D_{\mathrm{KL}}}

\newcommand{\indep}{\perp \!\!\! \perp}
\newcommand{\notindep}{\not\!\perp\!\!\!\perp}

\newcommand{\Normal}{\mathcal{N}}

\newcommand{\E}{\mathbb{E}}

\newcommand{\zhat}{\hat{\mathbf{z}}}

\newcommand{\bfx}{\mathbf{x}}

\newcommand{\bfz}{\mathbf{z}}

\newcommand{\bfy}{\mathbf{y}}

\newcommand{\bfv}{\mathbf{v}}

\newcommand{\cD}{\mathcal{D}}

\newcommand{\R}{\mathbb{R}}

\newcommand{\concat}{\mathrm{Concat}}

\newcommand{\RNN}{\mathrm{RNN}}

\newcommand{\EC}{\mathrm{EC}}
\newcommand{\EO}{\mathrm{EO}}

\title{InfoDPCCA: Information-Theoretic Dynamic Probabilistic Canonical Correlation Analysis}

\author[1,2]{\href{mailto:<t.sq@my.cityu.edu.hk>?Subject=Your UAI 2025 paper}{Shiqin Tang}} 
\author[3,4]{\href{mailto:<yusj9011@gmail.com>?Subject=Your UAI 2025 paper}{Shujian Yu}}
\affil[1]{
    Dept. of Data Science \\
    City University of Hong Kong\\
    Hong Kong}
\affil[2]{
    Inst. of Artificial Intelligence \\
    Peking University\\
    Beijing, China}
\affil[3]{
    Dept. of Computer Science,
    Vrije Universiteit Amsterdam,
    Amsterdam, Netherlands}
\affil[4]{
    Dept. of Physics and Technology \\
    UiT The Arctic University of Norway\\
    Troms\o¸, Norway}

\begin{document}
\maketitle

\begin{abstract}
Extracting meaningful latent representations from high-dimensional sequential data is a crucial challenge in machine learning, with applications spanning natural science and engineering.
We introduce InfoDPCCA, a dynamic probabilistic Canonical Correlation Analysis (CCA) framework designed to model two interdependent sequences of observations.
InfoDPCCA leverages a novel information-theoretic objective to extract a shared latent representation that captures the mutual structure between the data streams and balances representation compression and predictive sufficiency while also learning separate latent components that encode information specific to each sequence.
Unlike prior dynamic CCA models, such as DPCCA, our approach explicitly enforces the shared latent space to encode only the mutual information between the sequences, improving interpretability and robustness.
We further introduce a two-step training scheme to bridge the gap between information-theoretic representation learning and generative modeling, along with a residual connection mechanism to enhance training stability. Through experiments on synthetic and medical fMRI data, we demonstrate that InfoDPCCA excels as a tool for representation learning.
Code of InfoDPCCA is available at \url{https://github.com/marcusstang/InfoDPCCA}.
\end{abstract}

\section{Introduction}
\label{sec:intro}
Extracting meaningful latent representations from complex, high-dimensional sequential data is a fundamental challenge in machine learning, with broad applications across natural science and engineering. For instance, state-space models have been employed to uncover dynamic patterns in brain activity, aiding in the diagnosis of psychiatric disorders \citep{fmri16}, and to track functional connectivity changes in resting-state fMRI for cognitive impairment assessment \citep{fmri19}.

In this paper, we introduce InfoDPCCA, a dynamic probabilistic Canonical Correlation Analysis (CCA) framework designed to model two interdependent sequences of observations. Our approach extracts a shared latent representation that captures the mutual information between the two data streams while simultaneously learning distinct components unique to each sequence. By leveraging a novel information-theoretic objective and a two-step training scheme, InfoDPCCA encourages the shared latent states to encode only the mutual information between the observations and strike a balance between representation compression and predictive sufficiency. InfoDPCCA can be viewed both as a generative model and a stochastic representation learning method.

The paper is organized as follows:
In Section \ref{sec:related_works}, we review the recent development of CCA and the usage of Information Bottleneck in sequential modeling. Section \ref{sec:background} describes simpler probabilistic models that form the backbones of our approach.
In Section \ref{sec:dpcca}, we introduce Deep Dynamic Probabilistic CCA (D$^2$PCCA) \citep{d2pcca} as a baseline for comparison.
Section \ref{sec:fix1} compares the graphical models of D$^2$PCCA and InfoDPCCA and provides justifications for the change in design choice.
Section \ref{sec:fix2} identifies a key limitation of D$^2$PCCA and introduces an information-theoretic objective to mitigate this issue.
Section \ref{sec:training} outlines a two-step training scheme for InfoDPCCA.
Section \ref{sec:stab} proposes methods to improve the stability and efficiency of the training process.
Finally, in Section \ref{sec:exp}, we provide numerical simulations and real-world medical fMRI data to validate that InfoDPCCA fulfills its claims and excels as a tool for representation learning.

The major contributions of the paper include:
\begin{enumerate}
\item The proposal of a novel dynamic CCA model that improves upon its predecessor D$^2$PCCA.
\item A novel information-theoretic objective to extract shared latent states as well as its tractable variational bound.
\item A two-step training approach that bridge the gap between information-theoretic representation learning and generative modeling.
\item A novel residual connection proposed in Section \ref{sec:stab} to enhance the training stability.
\end{enumerate}

\section{Related Works}
\label{sec:related_works}
The method of using Information Bottleneck (IB) \citep{ib} to guide the training of sequential models is well-established.
\citet{gib} proposes Gaussian IB (GIB) that extends the IB principle to multivariate Gaussian variables, showing that GIB's optimal projections align with CCA's eigenstructure.
\citet{pfib09} and \citet{pfib15} apply IB to linear dynamical systems, focusing on model reduction and system realization.
\citet{nib1} extends IB to nonlinear encoding and decoding maps and proposes a non-parametric upper bound on the mutual information (MI) term.
\citet{ceb} discusses the possibility of extending its static IB framework to sequence learning.
\citet{cpic22} maximizes predictive information in the representation space and adapts InfoNCE's \citep{infoNCE} MI lower bound to achieve a low-variance approximation.
\citet{teb22} leverages IB to learn compressed latent representations while preserving relevant transfer entropy information, which quantifies the causal influence of a source process on the future states of a target process.
\citet{pfib24} further decomposes transfer entropy into its contributions from the source's past and the target's future using IB.
\citet{tlib24} applies IB to learn latent representations that retain essential system dynamics while discarding superfluous information, such as high-frequency noise, irrelevant features, and short-term dependencies.
Another notable extension is the Multi-view Information Bottleneck \citep{mvib}, which adapts IB to learn the representations of each view of the data that capture the shared information across multiple views.

CCA is a statistical method that explores the shared structure of paired datasets by finding linear projections that maximize their correlation.
\citet{ncca} extends CCA to nonlinear settings, making the latent representations more flexible and expressive.
Motivated by the intuition behind CCA, \citet{pcca_bach} and \citet{pcca_klami} propose probabilistic versions of CCA by modeling an underlying shared latent state that generates both observations.
\citet{dvcca} further extends probabilistic CCA to nonlinear observation models using amortized variational inference \citep{avi}.
Dynamic Probabilistic CCA (DPCCA) \citep{dpcca} is a dynamic extension of CCA that models the evolution of two variables simultaneously over time.
\citet{d2pcca} builds upon DPCCA by introducing nonlinear state transitions and emission models, making it more expressive for complex time-series data. We refer to both linear and nonlinear versions of the model as DPCCA.

\section{Background}
\label{sec:background}
In this section, we briefly introduce the static Variational Information Bottleneck (VIB) \citep{vib} and Conditional Entropy Bottleneck (CEB) \citep{ceb} models and propose a dynamic extension of VIB, which we name Dynamic VIB (DVIB). These models serve as the foundation for extending our approach to the more complex case of two observations.

Information Bottleneck (IB) \citep{ib} is a framework that seeks to optimize the trade-off between compressing the input $\bfx$ into a compact representation $\bfz$ while preserving relevant information about the target $\bfy$, with the goal of minimizing the mutual information between $\bfx$ and $\bfz$, $I(\bfx;\bfz)$, while maximizing the mutual information between $\bfz$ and $\bfy$, $I(\bfz;\bfy)$, i.e.
\begin{align}
\min\quad \beta I(\bfz;\bfx) - I(\bfz;\bfy), \label{eq:ib}
\end{align}
where $\beta$ is a hyperparameter controlling the contribution of the two terms.

VIB models the Markov chain $\bfy \leftrightarrow \bfx \leftrightarrow \bfz$, where $\bfz$ is a stochastic latent representation of $\bfx$. VIB considers an additional observed variable $\bfy$ such that the three variables follow the joint distribution $p(\bfx,\bfy, \bfz) = p(\bfx)p(\bfy|\bfx)q_\phi(\bfz|\bfx)$, where $\phi$ parameterizes the stochastic encoder.
VIB optimizes the IB \eqref{eq:ib} by replacing the mutual information terms with tractable variational bounds.

On the other hand, \citet{ceb} optimizes the CEB objective,
\begin{align}
\min\quad I(\bfz;\bfx|\bfy) - \lambda I(\bfz;\bfy), \label{eq:ceb}
\end{align}
where the conditional mutual information (CMI) is given by
\begin{align}
I(\bfz,\bfx|\bfy) &= I(\bfx;\bfz) - I(\bfz;\bfy). \label{eq:cmi}
\end{align}
The CEB objective can be rewritten in the same form as \eqref{eq:ib} since \eqref{eq:ceb} can be expressed as $I(\bfz;\bfx) - (1+\lambda)I(\bfz;\bfy)$.
However, unlike IB, the two terms in the CEB objective are disjoint, allowing CEB to achieve a tighter approximation through a better choice of the inference network.
Furthermore, by representing the IB as CEB, we can track the evolution of the conditional mutual information term, which can be driven to zero in an optimal case.

To obtain a lower bound on $I(\bfz;\bfy)$, both VIB and CEB require training a classifier network $p_\theta(\bfy|\bfz)$. Once training is complete, classification is performed as $p(\bfy|\bfx) = \E_{q_\phi(\bfz|\bfx)}[p_\theta(\bfy|\bfz)]$.
This contrasts with the Conditional Variational Autoencoder (CVAE) \citep{cvae} approach, where classification is given by $p(\bfy|\bfx) = \E_{q_\phi(\bfz|\bfx)}[p_\theta(\bfy|\bfz, \bfx)]$.
Unlike CVAE, which explicitly conditions on $\bfx$ in the decoder, VIB learns a stochastic encoder $q_\phi(\bfz|\bfx)$ that ensures the latent representation $\bfz$ is a sufficient statistic for predicting $\bfy$. By enforcing this constraint, the classification model becomes more robust to noise in $\bfx$. Furthermore, unlike CVAE, VIB does not require a separate inference network $q(\bfz|\bfx, \bfy)$ in training.

\begin{figure}
\begin{center}
\begin{tabular}{cc}
\includegraphics[trim={0cm 0cm 0cm 0cm}, clip, width=.23\textwidth]{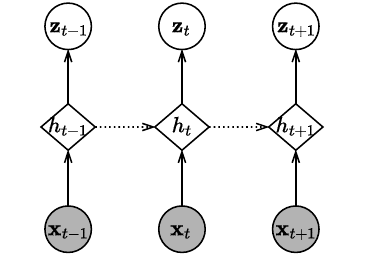} & \includegraphics[trim={0cm 0cm 0cm 0cm}, clip, width=.23\textwidth]{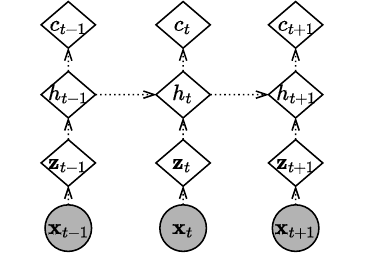}\\
(a) DVIB & (b) InfoNCE
\end{tabular}
\end{center}
\setlength{\abovecaptionskip}{0pt}
\caption{Graphical models for DVIB and InfoNCE. The shaded nodes denote observations, while the unshaded nodes denote latent states. The solid arrows denote stochastic connections, while the dotted arrows denote deterministic maps.}
\label{fig:dvib}
\end{figure}

We propose DVIB as a straightforward dynamic extension to VIB and a realization of the sequence learning approach depicted in the Equation (A4) of \citep{ceb}.
As illustrated in Figure \ref{fig:dvib}, DVIB can be regarded as a stochastic counterpart of InfoNCE in the case of one-step-ahead prediction;
specifically, the latent state of DVIB, $\bfz_t$, is comparable to $c_t$ of InfoNCE, which is computed by the hidden state $h_t$.
InfoNCE maximizes a lower bound on the mutual information between $c_t$ and the future states known as the noise contrastive loss.
By contrast, DVIB aims to learn a stochastic representation $\bfz_t$ that contains most information content of $\bfx_{1:t}$ in predicting $\bfx_{t+1}$.
To achieve this, we use RNN hidden states to encode historical input $\bfx_{1:t}$, i.e. $h_t = \RNN(h_{t-1}, \bfx_t)$.
The encoder $q_\phi(\bfz_t|\bfx_{1:t})$ and decoder $p_\theta(\bfx_{t+1}|\bfz_t)$ networks are trained by optimizing the IB in a sequential setting,
\begin{equation}
\begin{aligned}
\min\quad \sum_{t=1}^T\Big\{ \beta I(\bfz_t;\bfx_{1:t}) - I(\bfz_t;\bfx_{t+1})\Big\},
\end{aligned}\label{eq:dvib_obj1}
\end{equation}
with a tractable variational bound:
\begin{equation}
\begin{aligned}
\min\quad &\sum_{t=1}^T\Big\{ \langle \beta \log q_\phi(\bfz_t|h_t) - \log r(\bfz_t)\rangle \\
&\qquad\quad - \langle \log p_\theta(\bfx_{t+1}|\bfz_t)\rangle\Big\},
\end{aligned}\label{eq:dvib_obj2}
\end{equation}
\normalsize
where $r(\bfz_t)= \log \Normal(\bfz_t|0,I)$ can be chosen as the regularization term, and $\langle \cdot \rangle$ represents expectations taken with respect to $p(\bfx_{1:t+1})q_\phi(\bfz_t|h_{t})$. As will become evident later in the paper, DVIB's simplicity and the structure of its objective provide critical insights into InfoDPCCA's more general approach to modeling two sequences.


\section{Proposed Model: InfoDPCCA} 
\label{sec:infodpcca}
In this section, we first review DPCCA \citep{d2pcca} and show that the proposed InfoDPCCA model is well-motivated to ameliorate a key limitation of DPCCA in capturing mutual information.

Before introducing the proposed method, we first define the notation used throughout this paper.
We denote the two observations at time $t$ as $\bfx_t^1$ and $\bfx_t^2$.
The entire sequence of historical observations for variable $\bfx^i$ is represented as $\bfx_{1:t}^i$.
We use $\bfz_t^0$ to denote the latent states shared by both observations and use $\bfz_t^i$ to denote latent states private to observation $\bfx^i$.
Similar to the previous section, we use $h_t$ to denote RNN hidden states, with appropriate superscripts added to distinguish between different RNNs. We use $p_D$ to denote the empirical distribution of the dataset.

Following the notations of \citep{koller}, we say that $\bfx$ is conditionally independent from $\bfz$ given $\bfy$, denoted as $\bfx\indep \bfz|\bfy$, if either one of the following conditions hold:
\begin{align}
&p(\bfx,\bfz|\bfy) = p(\bfx|\bfy) p(\bfz|\bfy), \label{eq:indep1}\\
&p(\bfz|\bfx,\bfy) = p(\bfz|\bfy).  \label{eq:indep2}
\end{align}

\begin{figure}
\begin{center}
\begin{tabular}{ccc}
\includegraphics[trim={1cm 0cm 1.7cm 0cm}, width=.4\linewidth]{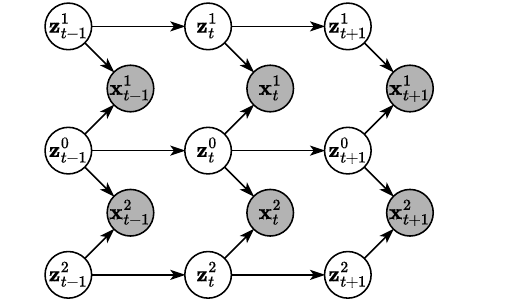}& &
\includegraphics[trim={1cm 0cm 1.7cm 0cm}, width=.413\linewidth]{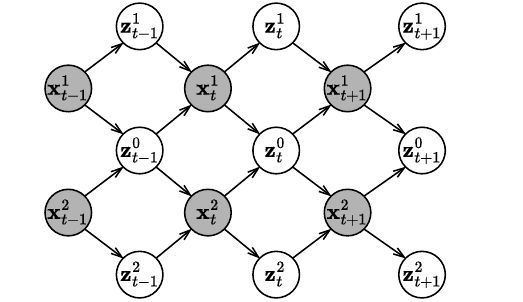}\\
(a) DPCCA && (b) InfoDPCCA
\end{tabular}
\end{center}
\setlength{\abovecaptionskip}{0pt}
\caption{Graphical models for DPCCA and a simplified version of InfoDPCCA. }
\label{fig:dpcca}
\end{figure}

\subsection{DPCCA: A Baseline for Comparison}
\label{sec:dpcca}

DPCCA is a latent state-space model where a shared latent variable $\bfz_t^0$ captures mutual dependencies between the two sequences, while unique latent variables $\bfz_z^1$ and $\bfz_z^2$ account for individual variations.
As is shown by the probabilistic graphical model in Figure \ref{fig:dpcca} (a), DPCCA satisfies the following set of (conditional) independence properties:
\begin{align}
z_{t+1}^i \indep z_{t+1}^j|\bfz_t^{0:2},\quad x_t^1 \indep x_t^2|\bfz_{t}^{0:2},\quad z_t^i \notindep z_t^j|\bfx_t^{1:2},\label{eq:dpcca_indep}
\end{align}
for $i,j \in \{0,1,2\}$.
Each recurrent segment of DPCCA has the state transition and emission models
\begin{align}
&p_\theta(\bfz_t^{0:2}|\bfz_{t-1}^{0:2}) = \prod_{i=0}^2 p_\theta(\bfz_t^{i}|\bfz_{t-1}^{i}),\\
&p_\theta(\bfx_t^{1:2}|\bfz_t^{0:2}) = p_\theta(x_t^1|z_t^1, z_t^0)p_\theta(x_t^2|z_t^2, z_t^0),\label{eq:dpcca_emission}
\end{align}
where $\theta$ denotes the parameters of the generative process.
DPCCA is trained by maximizing the variational lower bound (ELBO),
\begin{align}
\max_{\theta,\phi} \quad \E_{q_\phi(\bfz_{1:T}^{0:2}|\bfx_{1:T}^{1:2})}\left[\log \frac{p_\theta(\bfx_{1:T}^{1:2}, \bfz_{1:T}^{0:2})}{q_\phi(\bfz_{1:T}^{0:2}|\bfx_{1:T}^{1:2})}\right],
\label{eq:dpcca_elbo}
\end{align}
where the inference network $q_\phi$ has the form
\begin{align}
q_\phi(\bfz_{1:T}^{0:2}|\bfx_{1:T}^{1:2}) &= \prod_{t=1}^T q(\bfz_t^{0:2}|\bfz_{t-1}^{0:2}, \bfx_{t:T}^{1:2}).
\label{eq:dpcca_elbo}
\end{align}

\subsection{Comparison of Graphical Models}
\label{sec:fix1}

\begin{figure}
\begin{center}
\begin{tabular}{ccc}
\includegraphics[trim={1cm 0cm 1.8cm 0cm}, width=.42\linewidth]{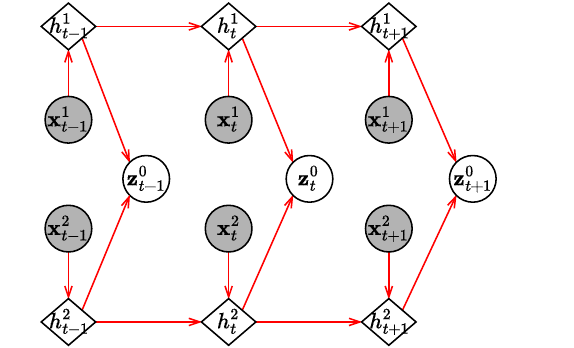}& &
\includegraphics[trim={1cm 0cm 1.8cm 0cm}, width=.42\linewidth]{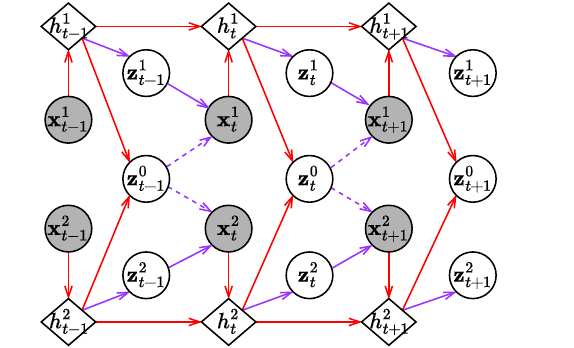}\\
(a) Step I && (b) Step II
\end{tabular}
\end{center}
\setlength{\abovecaptionskip}{0pt}
\caption{Graphical models in the two-step training of InfoDPCCA. The diamond nodes denote RNN hidden states. We use solid red arrows to denote the connections learned in step I, and solid purple arrows to denote the connections trained in step II, and dashed purple arrows to denote the residual connections from step I.}
\label{fig:infodpcca}
\end{figure}

For ease of comparison, we illustrate the graphical model of a simplified version of InfoDPCCA in Figure \ref{fig:dpcca} (b), which assumes that the current hidden state is conditionally independent of previous observations given the current observations. However, in practice, we consider a more general case where each hidden state depends on all previous observations. The graphical model for this general case of InfoDPCCA is shown in Figure \ref{fig:infodpcca} (b).

It can be verified that, similar to DPCCA \eqref{eq:dpcca_indep}, InfoDPCCA satisfies the following conditional independence properties:
\begin{align}
z_{t+1}^i \indep z_{t+1}^j|\bfx_{t+1}^{1:2},\quad x_t^1 \indep x_t^2|\bfz_{t-1}^{0:2},\quad z_t^i \notindep z_t^j|\bfx_{t+1}^{1:2}.\label{eq:infodpcca_indep}
\end{align}

A major difference between the two models is that InfoDPCCA assumes a conditional independence property for the distribution of its  latent states, which we refer to as serial independence, defined as
\begin{align}
z_t^i \indep z_{t-n}^j|\bfx_{1:T}^{1:2},\quad \forall i,j \in \{0,1, 2\}.
\label{eq:serial_indep}
\end{align}
This assumption implies that given all the observations, the private latent states of different observations at different time steps are conditionally independent. The serial independence of InfoDPCCA can be established by
\begin{align}
p(z_t^i, z_{t-n}^j|\bfx_{1:T}^{1:2}) = p(z_t^i|\bfx_{1:t+1}^{1:2})p(z_{t-n}^j|\bfx_{1:t-n+1}^{1:2}),
\end{align}
according to the criteria \eqref{eq:indep1}.
These properties hold for both the simplified and general versions of InfoDPCCA.
On the other hand, in the case of DPCCA,  we can prove $z_{t-1}^1 \notindep z_t^2|\bfx_{1:T}^{1:2}$ using \eqref{eq:indep1},
\smaller
\begin{align*}
&p(z_{t-1}^1, z_t^2|\bfx_{1:T}^{1:2}) = \sum_{z_{t-1}^0, z_{t-1}^2} p(z_{t-1}^1, z_{t-1}^0, z_{t-1}^2,z_t^2|\bfx_{1:T}^{1:2})\\
&\qquad \qquad = p(z_{t-1}^1|\bfx_{1:T}^{1:2})  \sum_{z_{t-1}^0, z_{t-1}^2} p(z_{t-1}^0, z_{t-1}^2,z_t^2|\bfx_{1:T}^{1:2}, z_{t-1}^1)\\
&\qquad \qquad\neq p(z_{t-1}^1|\bfx_{1:T}^{1:2}) p(z_{t}^2|\bfx_{1:T}^{1:2}).
\end{align*}
\normalsize

Admittedly, the graphical model of DPCCA may seem more intuitive; however, we believe that the mesh structure of InfoDPCCA is necessary to facilitate information-theoretic representation learning. Generative models like VAEs or DPCCA assume that the observations are generated by the latent variables ($\bfz_t \rightarrow \bfx_t$), while representation learning methods that apply information-theoretic principles like VIB and Multi-view IB treat the latent variables as disentangled representations of the observations ($\bfx_t \rightarrow \bfz_t$). InfoDPCCA falls into the latter category, and its objective function is analyzed in detail in Section \ref{sec:fix2}.

Furthermore, the serial independence assumption aligns with other established methods when applied to, for example, industrial processes or single-cell biomedical sequences.
LaVAR-CCA \citep{qin} combines contemporaneous orthogonality, where latent variables are uncorrelated at each time step, with a vector autoregressive model that captures temporal dependencies.
A conceptually similar approach to InfoDPCCA appears in Tilted-CCA \citep{lin}, which, while applied to non-sequential single-cell data, uses orthogonality constraints to separate shared and distinct information across modalities.

\subsection{Information-Theoretic Objective}
\label{sec:fix2}
The shared latent states $\bfz_t^0$ of DPCCA have an uncertain meaning.
Beyond the structural design of the generative model, there is no guarantee that the extracted shared latent states exclusively capture the mutual information between the two observations, nor is the model explicitly trained to enforce this property.
For instance, if the dimensionality of $\bfz_t^0$ is relatively small, the private latent states $\bfz_t^1$ and $\bfz_t^2$ may independently encode the dynamics of $\bfx_t^1$ and $\bfx_t^2$, respectively, leading to a low reconstruction loss; in such case $\bfz_t^0$ may primarily encode residual dynamics that lack meaningful shared structure.

\begin{figure}
\begin{center}
\begin{tabular}{c}
\includegraphics[clip, width=.98\linewidth]{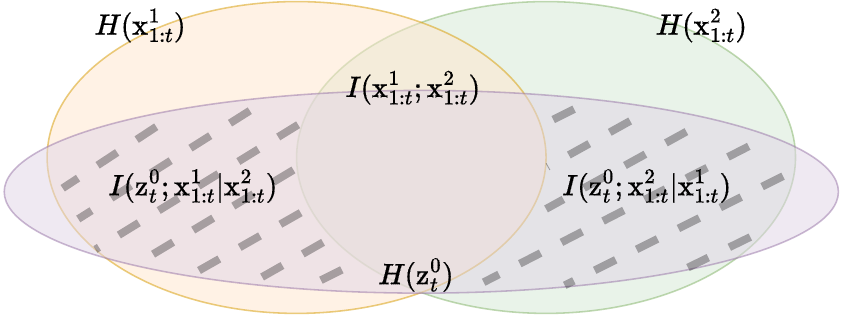}
\end{tabular}
\end{center}
\setlength{\abovecaptionskip}{0pt}
\caption{
Venn diagram illustrating the relationship between the shared latent state $\bfz_t^0$ and the observations $\bfx_{1:t}^{1:2}$.
The shaded regions represent the regularization term in \eqref{eq:infodpcca_obj1}, which InfoDPCCA aims to minimize.
}
\label{fig:venn_z0}
\end{figure}

To remedy this, InfoDPCCA uses an information-theoretic objective to encourage the shared latent states $\bfz_t^0$ to exclusively capture the mutual information between the two observations.
We extract latent states $\bfz_{t}^{0}$ as the stochastic embeddings of $\bfx_{1:t}^{1:2}$ by optimizing the following objective,
\begin{equation}
\begin{aligned}
\min\quad &\sum_{t=1}^T \Big\{\underbrace{\alpha I(\bfz_t^{0}; \bfx_{1:t}^{1:2}) - I(\bfz_t^{0}; \bfx_{t+1}^{1:2})}_{\text{IB term}}\\
& + \beta \underbrace{\big(I(\bfz_t^0;\bfx_{1:t}^1|\bfx_{1:t}^2)+ I(\bfz_t^0;\bfx_{1:t}^2|\bfx_{1:t}^1) \big)}_{\text{regularization term}} \Big\},
\end{aligned}\label{eq:infodpcca_obj1}
\end{equation}
where the IB term is similar to \eqref{eq:ib} and \eqref{eq:dvib_obj1} and the regularization term discourage $\bfz_t^0$ to learn sequence-specific information as illustrated in Figure \ref{fig:venn_z0}.

As depicted in Figure \ref{fig:infodpcca} (a), our framework assumes
that $\bfz_t^0$ is generated by $\bfx_{1:t}^{1:2}$ via a stochastic map,
\begin{align}
p(\bfz_t^{0}|\bfx_{1:t}^{1:2}) &= q_0^{1:2}(\bfz_t^{0}|\bfx_{1:t}^{1:2})= q_0^{1:2}(\bfz_t^{0}|h_{t}^1, h_{t}^2),
\end{align}
where $h_t^i = \RNN(h_t^i, \bfx_t^i)$ for $i = 1,2$.
We also assume that the emission model can be factored as
\begin{align}
p(\bfx_{t+1}^{1:2}|\bfz_{t}^{0}) &= p_1^{0}(\bfx_{t+1}^{1}|\bfz_{t}^{0}) p_2^{0}(\bfx_{t+1}^{2}|\bfz_{t}^{0}).
\end{align}
Following the idea of VIB, we have the following bounds
\begin{align}
I(\bfz_t^{0}; \bfx_{1:t}^{1:2}) &\leq \langle \log q_0^{1:2}(\bfz_t^{0}|\bfx_{1:t}^{1:2}) -\log r(\bfz_t^{0}) \rangle, \label{eq:eq19}\\
I(\bfz_t^{0}; \bfx_{t+1}^{1:2}) &\geq \langle \log p_1^{0}(\bfx_{t+1}^{1}|\bfz_{t}^{0}) p_2^{0}(\bfx_{t+1}^{2}|\bfz_{t}^{0}) \rangle + \text{const.}, \label{eq:eq20}
\end{align}
where $\langle \cdot \rangle$ represent expectations taken with respect to  $p_D(\bfx_{1:t+1}^{1:2})q_0^{1:2}(\bfz_t^0|\bfx_{1:t}^{1:2})$.

To further simplify the regularization term in \eqref{eq:infodpcca_obj1}, we temporarily simplify the notations by removing the superscripts of each variable and using their original superscripts as their subscripts, e.g. $\bfx_{1:t}^{1:2} \rightarrow \bfx_{1:2}$. Firstly, we have the equality
\smaller
\begin{equation}
\begin{aligned}
&I(\bfz_0;\bfx_1|\bfx_2) + I(\bfz_0;\bfx_2|\bfx_1) \\
=\, &\big(H(\bfz_0;\bfx_2) - H(\bfz_0; \bfx_{1:2})\big) + \big(H(\bfz_0;\bfx_1)-H(\bfz_0; \bfx_{1:2})\big)\\
=\, &\big(I(\bfz_0; \bfx_{1:2}) - I(\bfz_0;\bfx_2)\big) + \big(I(\bfz_0; \bfx_{1:2})- I(\bfz_0;\bfx_1)\big).
\end{aligned}
\end{equation}
\normalsize
Furthermore, we can establish a variational bound
\begin{equation}
\begin{aligned}
&I(\bfz_0; \bfx_{1:2}) - I(\bfz_0;\bfx_1)\\
=\,& \iint p(\bfz_0, \bfx_{1:2}) \log \frac{p(\bfz_0| \bfx_{1:2})}{\cancel{p(\bfz_0)}} \d \bfz_0 \d \bfx_{1:2} \\
&\qquad - \iint p(\bfz_0, \bfx_{1}) \log \frac{p(\bfz_0| \bfx_{1})}{\cancel{p(\bfz_0)}} \d \bfz_0 \d \bfx_{1}\\
\overset{(a)}{\leq}\,& \E_{p(\bfz_0, \bfx_{1:2})} [\log q_0^{1:2}(\bfz_0| \bfx_{1:2})] - \E_{p(\bfz_0, \bfx_{1})} [\log q_0^1(\bfz_0| \bfx_{1})]\\
\approx\,& \langle\log q_0^{1:2}(\bfz_0|\bfx_{1:2})\rangle - \langle\log q_0^{1}(\bfz_0|\bfx_{1})\rangle,
\end{aligned}\label{eq:var_bound}
\end{equation}
where the inequality in (a) holds due to the modeling assumption that $p(\bfz_0| \bfx_{1:2})=q_0^{1:2}(\bfz_0| \bfx_{1:2})$ and the Gibbs' inequality
\begin{equation}
\begin{aligned}
0 \leq\,  & \E_{p(\bfx_1)}[\KL(p(\bfz_0| \bfx_{1}) \| q_0^1(\bfz_0| \bfx_{1}))]\\
=\, &\E_{p(\bfz_0, \bfx_{1})} [\log p(\bfz_0| \bfx_{1})] - \E_{p(\bfz_0, \bfx_{1})} [\log q_0^1(\bfz_0| \bfx_{1})],
\end{aligned}\label{eq:gibbs}
\end{equation}
and the expectations are defined as
\smaller
\begin{equation}
\begin{aligned}
\langle\log q_0^{1:2}(\bfz_0|\bfx_{1:2})\rangle&= \E_{p_D(\bfx_{1:t}^{1:2}) q_0^{1:2}(\bfz_t^0|\bfx_{1:t}^{1:2})}[\log q_0^{1:2}(\bfz_t^0|\bfx_{1:t}^{1:2})], \\
\langle\log q_0^{1}(\bfz_0|\bfx_{1})\rangle &= \E_{p_D(\bfx_{1:t}^{1:2})q_0^{1:2}(\bfz_t^0|\bfx_{1:t}^{1:2})}[\log q_0^{1}(\bfz_t^0|\bfx_{1:t}^{1})].
\end{aligned}\label{eq:exp_def}
\end{equation}
\normalsize
Similarly, we obtain the variational bound,
\begin{equation}
\begin{aligned}
&I(\bfz_0; \bfx_{1:2}) - I(\bfz_0;\bfx_2) \\
\leq \,&\langle\log q_0^{1:2}(\bfz_t^0|\bfx_{1:t}^{1:2})\rangle - \langle\log q_0^{2}(\bfz_t^0|\bfx_{1:t}^{2})\rangle.
\end{aligned}
\end{equation}


In summary, we solve for encoder $q_0^{1:2}$, $q_0^{1}$, $q_0^{2}$, and decoder $p_1^0$, $p_2^0$ networks by optimizing the following objective,
\begin{equation}
\begin{aligned}
\min\quad  & (\alpha+2\beta)\langle \log q_0^{1:2}(\bfz_t^0|\bfx_{1:t}^{1:2})\rangle - \alpha \langle r(\bfz_t^{0})\rangle \\
&\quad - \langle \log p_1^0(\bfx_{t+1}^{1}|\bfz_{t}^{0}) \rangle -\langle\log p_2^0(\bfx_{t+1}^{2}|\bfz_{t}^{0}) \rangle\\
&\quad - \beta \langle\log q_0^{1}(\bfz_t^0|\bfx_{1:t}^{1})\rangle - \beta \langle\log q_0^{2}(\bfz_t^0|\bfx_{1:t}^{2})\rangle,
\end{aligned}\label{eq:infodpcca_obj2}
\end{equation}
where we use the bounds developed in Equations \eqref{eq:eq19} and \eqref{eq:eq20}.
The extracted shared latent states $\bfz_{t}^0$,
as formulated above, achieve a balance between minimizing overall information content, ensuring they capture only the shared components between the two observations, and retaining sufficient information to predict the next observation.

Unfortunately, there's no closed-form solution to pick $\alpha$ and $\beta$. One may use cross-validation to set the hyperparameters. By default, we can set $\alpha = 1$ and $\beta = 0.1$.

\begin{rmrk} While the Gibbs' inequality \eqref{eq:gibbs} holds for any choice of distribution over $\bfz_0$, the tightness of the variational bound \eqref{eq:var_bound} depends on how well $q_0^1(\bfz_0|\bfx_1)$ approximates the true posterior $p(\bfz_0|\bfx_1)$.
In fact, the training objective \eqref{eq:infodpcca_obj2} is designed to encourage this match by including the term $-\beta\langle\log q_0^{1}(\bfz_0|\bfx_{1})\rangle$ (as defined in \eqref{eq:exp_def}).
Minimizing this term drives $q_0^{1}(\bfz_0|\bfx_{1})$ to align with $q_0^{1:2}(\bfz_0|\bfx_{1:2})$, which serves as an approximation to $p(\bfz_0|\bfx_1)$.
To see why, note that
\smaller
\begin{equation}
\begin{aligned}
p(\bfz_0|\bfx_1) &= \int p(\bfz_0, \bfx_2|\bfx_1) \d \bfx_2 = \int p(\bfx_2|\bfx_1) q_0^{1:2}(\bfz_0| \bfx_{1:2}) \d \bfx_2 \\
&\approx \int p_D(\bfx_2|\bfx_1) q_0^{1:2}(\bfz_0| \bfx_{1:2}) \d \bfx_2 = q_0^{1:2}(\bfz_0| \bfx_{1:2}).
\end{aligned}
\end{equation}
\normalsize
\end{rmrk}

\subsection{Two-Step Training Scheme}
\label{sec:training}

Instead of training the model in a single step, we propose a novel two-step approach for InfoDPCCA. In the first step, we extract the shared latent states that capture the mutual information between the two observations as mentioned in Section \ref{sec:fix2}. In the second step, we train the entire generative model, including the private latent states, while keeping the parameters learned in the first step fixed.


We use variational inference to train the remaining components of the generative model, specifically the private latent states and their associated stochastic mappings. The goal is for $\bfz_t^0$ to capture the shared global dynamics of the system, while $\bfz_t^1$ and $\bfz_t^2$ encode local variations and residual fluctuations. As shown in Figure \ref{fig:infodpcca} (b), solid red arrows indicate the connections trained in Step I, while solid purple arrows represent the connections trained in Step II.

With the parameters of the encoder network $q_0^{1:2}$ fixed, we then solve for $\theta$ and $\phi$ in
\begin{align}
\max\quad \E_{q_\phi(\bfz_{1:T}^{0:2}|\bfx_{1:T}^{1:2})}\Big[\log \frac{p_\theta(\bfz_{1:T}^{0:2},\bfx_{1:T}^{1:2})}{q_\phi(\bfz_{1:T}^{0:2}|\bfx_{1:T}^{1:2})}\Big],\label{eq:infodpcca_elbo}
\end{align}
where
\begin{equation}
\begin{aligned}
p_\theta(\bfz_{1:T}^{0:2},\bfx_{1:T}^{1:2}) &= \prod_{t=1}^T q_0^{1:2}(\bfz_{t}^0|\bfx_{1:t}^{1:2})p_\theta(\bfz_{t}^1|\bfx_{1:t}^{1})p_\theta(\bfz_{t}^2|\bfx_{1:t}^{2})\\
&\qquad\qquad p_\theta(\bfx_{t+1}^1|\bfz_{t}^{0,1}) p_\theta(\bfx_{t+1}^2|\bfz_{t}^{0,2}),\\
q_\phi(\bfz_{1:T}^{0:2}|\bfx_{1:T}^{1:2}) &= \prod_{t=1}^T q_\phi(\bfz_t^{0:2}|\bfx_{1:t+1}^{1:2}).
\end{aligned}\label{eq:factored}
\end{equation}
Note that we design the approximate posterior in a way that its factorization aligns in form with the true posterior factorization,
\begin{align}
p(\bfz_{1:T}^{0:2}|\bfx_{1:T}^{1:2}) &= \prod_{t=1}^T p(\bfz_t^{0:2}|\bfx_{1:t+1}^{1:2}).
\end{align}
In summary, the generative process is given by $p_\theta(\bfz_{1:T},\bfx_{1:T}^{1:2})$, which can be used for sequence generation and prediction. Meanwhile, the inference model $q_\phi(\bfz_{1:T}|\bfx_{1:T}^{1:2})$ is used to extract the latent states $\bfz_t^{0:2}$ for down-stream tasks.


\begin{figure}
\begin{center}
\begin{tabular}{@{}cc@{}}
\includegraphics[trim={0cm 0cm 0cm 0cm}, width=.35\linewidth]{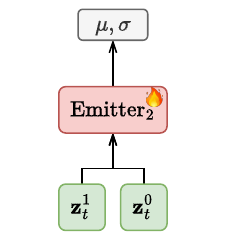} &\includegraphics[trim={0cm 0cm 0cm 0cm}, width=.53\linewidth]{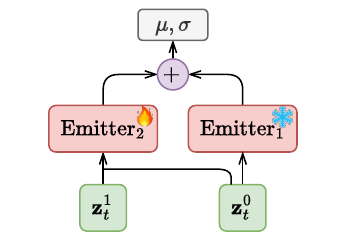}\\
(a) New Emitter & (b) Residual Connection
\end{tabular}
\end{center}
\setlength{\abovecaptionskip}{0pt}
\caption{Two neural network structures for InfoDPCCA's emission model. (a) shows a new emitter network trained from scratch, with trainable parameters indicated by fire icons. (b) reuses the emitter network from Step I via a residual connection, with fixed parameters indicated by ice icons.}
\label{fig:emitter}
\end{figure}


\subsection{Improving Stability \& Efficiency}
\label{sec:stab}

If the goal is for $\bfz_t^0$ to encode the primary or slow dynamics of the system while $\bfz_t^{1}$ and $\bfz_t^{2}$ capture the residual structures, is it optimal to reuse only the stochastic mapping $q_0^{1:2}(\bfz_t^0|\bfx_{1:t}^{1:2})$, as indicated by the purple dashed arrows in Figure \ref{fig:emitter} (b), while ignoring other inference networks trained in Step I?
In this section, we explore several approaches to leveraging the learned structures from Step I. In practice, these methods enhance both stability and efficiency (reduction in the total amount of parameters) in training.

Similar to the other deep state space models \citep{dvae, my_dssm, dmm}, we define the decoder network $p(\bfx|\bfz)$ as a Gaussian distribution, where the mean and variance are parameterized by a neural network referred to as the emitter:
\begin{align}
p_\theta(\bfx|\bfz) = \Normal(\mu_\theta(\bfz), \sigma_\theta(\bfz)^2),
\end{align}
where $(\mu_\theta(\bfz), \sigma_\theta(\bfz)) = \mathrm{Emitter}_\theta(\bfz)$.
In practice, we implement the emitter using a multi-layer perceptron (MLP). For example, we denote a three-layer MLP with ReLU activations in the hidden layers and a Sigmoid activation in the output layer as:
\begin{align}
\mathrm{Emitter}(\bfz) = \MLP(\bfz, \ReLU, \ReLU, \mathrm{Sigmoid}).
\end{align}


\textbf{Residual connections}. To optimize the information bottleneck objective \eqref{eq:infodpcca_obj1} in step I, we obtain a byproduct $p_1(\bfx_{t+1}^1|\bfz_t^0)$ associated with $\mathrm{Emitter_1}$.
This raises the question of whether or not $\mathrm{Emitter_1}$ can be reused in Step II.
Inspired by ResNet \citep{resnet}, we propose two designs for the decoder network $p_\theta(\bfx_{t+1}^1|\bfz_t^0, \bfz_t^1)$, referred to as $\mathrm{Emitter_2}$.
Figure \ref{fig:emitter} (a) illustrates a baseline approach where a separate emitter is trained from scratch:
\begin{align}
\mathrm{Emitter}_2(\bfz_t^{0:1}) = \MLP([\bfz_t^{0:1}], \ReLU, \ReLU, \mathrm{Sigmoid}). \label{eq:emittor_scratch}
\end{align}
In contrast, Figure \ref{fig:emitter} (b) presents an alternative approach where a residual connection from $\mathrm{Emitter_1}(\bfz_t^0)$ is incorporated into $\mathrm{Emitter_2}(\bfz_t^0, \bfz_t^1)$. To adaptively control the information flow from $\mathrm{Emitter_1}$, we introduce gating units \citep{cho2014gru, dmm}, leading to the following parameterization:
\begin{equation}
\begin{aligned}
&[\mu_1, \sigma_1] = \mathrm{Emitter}_2(\bfz_t^{0}),\\
&[\tilde{\mu}_2, \sigma_2] = \MLP([\bfz_t^{0},\bfz_t^{1}], \ReLU, \ReLU, \mathrm{Sigmoid}),\\
&g_t = \MLP([\bfz_t^{0}, \bfz_t^{1}], \ReLU, \mathrm{Sigmoid}),\, \smaller{\textit{(gating units)}}\\
&\mu_2 = (1-g_t) \odot \mu_1 + g_t \odot \tilde{\mu}_2,
\end{aligned}\label{eq:res_con}
\end{equation}
where $\odot$ denotes element-wise product.
As shown in \eqref{eq:res_con}, only $\mu_1$ is directly used to compute $\mu_2$, but we retain the flexibility to reuse both $\mu_1$ and $\sigma_1$. The parameters of $\mathrm{Emitter_1}$ remain fixed in Step II.
The use of residual connections can be particularly beneficial when $\bfz_t^0$ has a strong influence on the output, allowing the contribution of $\bfz_t^1$ to be adjusted adaptively via the residual connection alone.

\begin{algorithm}[!t]
\caption{Two-Step Training for \texttt{InfoDPCCA}}\label{alg:alg}
\KwIn{Observed paired sequences \( \{\bfx^1_{1:T}, \bfx^2_{1:T}\} \); Hyperparameters \( \alpha, \beta \) from \eqref{eq:infodpcca_obj1}.}
\textbf{// Step I: Optimizing info-theoretic objective}\;
Initialize parameters for inference networks $q_0^{1:2}$, $q_0^{1}$, $q_0^{2}$, $p_1^{0}$, $p_2^{0}$, and RNNs $d^1$ and $d^2$\;
\Repeat{convergence}{
$h_{1:T}^1 \leftarrow d^1(\bfx_{1:T}^1), \quad h_{1:T}^2 \leftarrow d^2(\bfx_{1:T}^2)$\;
Sample $\bfz_t^0 \sim q_0^{1:2}(\bfz_t^{0}|h_{t}^1, h_{t}^2)$\;
Compute the objective \eqref{eq:infodpcca_obj2}\;
Update \( q_0^{1:2}, q_0^{1}, q_0^{2}, p_1^{0}, p_2^{0}, d^1, d^2 \) via stochastic gradient descent algorithm\;
}
\textbf{// Step II: Full generative model training}\;
\uIf{\texttt{residual\_connection}}{
    Construct $\mathrm{Emitter}_2$ in \eqref{eq:res_con} using $p_1^0$ and $p_2^0$\;
  }
  \Else{
    Construct $\mathrm{Emitter}_2$ in \eqref{eq:emittor_scratch}\;
  }
\Repeat{convergence}{
\uIf{\texttt{reuse\_RNN}}{
    $h_{1:T}^1 \leftarrow d^1(\bfx_{1:T}^1), \quad h_{1:T}^2 \leftarrow d^2(\bfx_{1:T}^2)$\;
    Sample $\bfz_{t}^{0:2}$ following procedures \eqref{eq:info_post_sample}\;
  }
  \Else{
    Sample $\bfz_{t}^{0:2} \sim q_\phi(\bfz_{t}^{0:2}|\bfx_{1:t+1}^{1:2})$ in \eqref{eq:factored}\;
  }
Compute the objective \eqref{eq:infodpcca_elbo}\;
Update \( \theta, \phi \) via stochastic gradient descent\;
}
$h_{1:T}^1 \leftarrow d^1(\bfx_{1:T}^1), \quad h_{1:T}^2 \leftarrow d^2(\bfx_{1:T}^2)$\;
$\mu_t \leftarrow W_\mu[h_{t+1}^1, h_{t+1}^2] + b_\mu$\;
\KwOut{Extracted latent states $\bfz_{1:T}^{0:2} \leftarrow \mu_{1:T}$}
\end{algorithm}

\textbf{Inference network design}.
The optimization of the ELBO \eqref{eq:infodpcca_elbo} in Step II requires a separate inference network $q_\phi(\bfz_{1:T}^{0:2}|\bfx_{1:T}^{1:2})$, which introduces redundancy in the model parameters.
Inspired by Variational RNN (VRNN) \citep{vrnn}, we can mitigate this redundancy by reusing the RNN hidden states from Step I.

Specifically, we adopt the following parameterization, which results in $q_\phi(\bfz_{t}^{0:2}|\bfx_{1:t+1}^{1:2}) = \Normal(\bfz_{t}^{0:2}|\mu_t, \mathrm{diag}(\sigma_t^2))$ as in \eqref{eq:factored}:
\begin{equation}
\begin{aligned}
&h_{1:T}^1 = d^1(\bfx_{1:T}^1), \quad h_{1:T}^2 = d^2(\bfx_{1:T}^2),\\
&\mu_t = W_\mu[h_{t+1}^1, h_{t+1}^2] + b_\mu,\\
&\sigma_t = \mathrm{Softplus}(W_\sigma[h_{t+1}^1, h_{t+1}^2] + b_\mu),
\end{aligned}\label{eq:info_post_sample}
\end{equation}
where $h_{t+1}^1$ and $h_{t+1}^2$ are RNN hidden states, as represented by the diamond-shaped nodes in Figure \ref{fig:infodpcca}, which together encode information from $\bfx_{1:t+1}^{1:2}$.

By applying residual connections and reusing the generative RNN in Step II, empirical evidence shows that this approach leads to improved training stability and a reduction in parameter size. A high-level algorithmic summary is provided in Algorithm \ref{alg:alg}.



\section{Experiments}
\label{sec:exp}

In this section, we present a series of experiments to evaluate the performance of InfoDPCCA. First, we conduct numerical simulations using synthetic data generated by the HÃ©non map to assess the model's ability to recover latent structures. Then, we apply InfoDPCCA to real-world medical fMRI data, where we aim to demonstrate its effectiveness in extracting shared latent dynamics across different neural states and patient groups. These experiments highlight the advantages of InfoDPCCA in both synthetic and real-world applications.

\subsection{Numerical Simulation}
\label{sec:exp_sim}

Since there is no known creteria for validating whether a model extracts the shared latent variables $\bfz_t^0$ only encodes the mutual information between the observations $I(\bfx_{1:t}^1; \bfx_{1:t}^2)$,  we designed a synthetic dataset with known latent states and developed a metric that measures how similar the extracted shared latent states are to the ground truth.

The dataset is generated using the HÃ©non map, a two-dimensional chaotic system defined by \( x_{t+1} = 1 - 1.4 x_t^2 + y_t \) and \( y_{t+1} = 0.3 x_t \). Each sequence starts from a random initial condition \( (x_0, y_0) \) and evolves for \( T \) time steps. The latent states are linearly mapped to observed variables \( \mathbf{x}_t \in \mathbb{R}^{d_x} \) and \( \mathbf{y}_t \in \mathbb{R}^{d_y} \) using randomly initialized matrices, with added Gaussian noise. The dataset consists of \( N \) sequences and is split into 80\% training and 20\% testing.

We can calculate the Pearson correlation coefficient between the true latent states $\bfz$ and the extracted ones $\zhat$.
Let $\bfz_t,\zhat_t \in \R^D$.
We define global-mean correlation metric as
\footnotesize
$$\rho_{jk} := \frac{\sum_{n=1}^N\sum_{t=1}^T (z_{ntj} - \overline{z_{j}})(\hat{z}_{ntk} - \overline{\hat{z}_{k}})}{\sqrt{\sum_{n=1}^N\sum_{t=1}^T (z_{ntj} - \overline{z_{j}})^2}\sqrt{\sum_{n=1}^N\sum_{t=1}^T (\hat{z}_{ntk} - \overline{\hat{z}_{k}})^2}},$$
\normalsize
where
$$\overline{z_{j}} = \frac{1}{NT}\sum_{n=1}^N\sum_{t=1}^T z_{ntj},\quad \overline{\hat{z}_{k}} = \frac{1}{NT}\sum_{n=1}^N\sum_{t=1}^T \hat{z}_{ntk}.$$
\normalsize
Finally, we report the sum of maximum correlations
\begin{align}
\hat{\rho} = \frac{1}{D}\sum_{j=1}^D \max_{k\in [D]} |\rho_{jk}|.\label{eq:corr}
\end{align}
\normalsize
Note that metrics like RMSE would not work here because, for instance, a model that extracts the latent states exactly like the ground truth but with an opposite sign would fail the RMSE test.

\begin{figure}
\begin{center}
\begin{tabular}{c}
\includegraphics[clip, width=.98\linewidth]{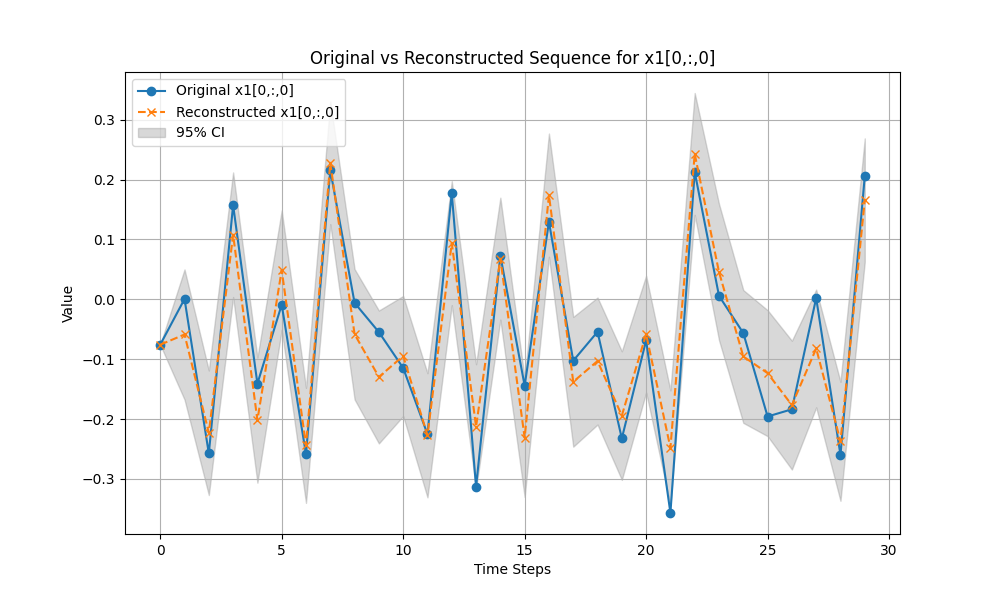}
\end{tabular}
\end{center}
\setlength{\abovecaptionskip}{0pt}
\caption{
Reconstruction on an HÃ©non map using InfoDPCCA.
}
\label{fig:recon}
\end{figure}

A visual inspection on the reconstruction can be seen in Figure \ref{fig:recon}. As can be seen, the reconstructed mean closely monitors the actual data, and every data point is confined within the predicted confidence interval.

The step I of InfoDPCCA is an info-theoretic representation learning method, while step II is a standard generative model that creates a separate inference network and maximizes an ELBO objective.
The global-mean correlation metric for InfoDPCCA with only step II is 65\% and the full two-step InfoDPCCA achieves a score of 72\%, which demonstrates the effectiveness of step I in capturing the underlying dynamics.

\subsection{Medical Data}
\label{sec:exp_real}

We evaluate our approach using three publicly available datasets, which consist of resting - state functional magnetic resonance imaging (rs-fMRI) data corresponding to different neural states or mental disorders. Our hypothesis is that the rs-fMRI data and its corresponding blood oxygen level - dependent (BOLD) signals can be modelled as a nonlinear dynamic system. Moreover, we assume that the common dynamics or the main driving forces corresponding to different neural states are distinguishable, and these could be further exploited as potential biomarkers for future neuroscience research.

The first dataset is the eyes closed and eyes open (ECEO) dataset \citep{eceo}. It contains rs-fMRI data of 48 college students (22 females) aged between 19 and 31 years. These students were scanned in both the eyes open and eyes closed states, yielding a total of 96 samples. The task is to differentiate between the two states using the rs-fMRI data. The second dataset is from the Alzheimer's Disease Neuroimaging Initiative (ADNI) database \citep{adni}. Here, the objective is to distinguish the Alzheimer's Disease (AD) group, consisting of 37 patients, from the normal control (NC) subjects, with a total of 37 individuals.
The third dataset is from the Autism Brain Imaging Data Exchange I (ABIDE - I) \citep{abide}. This dataset contains rs-fMRI data of 184 patients with Autism Spectrum Disorder (ASD) and 571 typically developed (TD) individuals, collected from 17 sites. In this paper, we utilize data collected from NYU and choose 40 ASD and 40 TD samples.
For each dataset, the Automated Anatomical Labeling (AAL) template was employed to extract the region-of-interest (ROI)-averaged time series from 116 ROIs. In other words, the rs-fMRI data was transformed into a 116-dimensional time series.

We utilize our proposed InfoDPCCA to extract shared latent states from pairs of rs-fMRI data within the same group (for example, both are from the eye-closed state or both are from schizophrenia patients).
Since there is not yet a criterion to verify the claim that the extracted $\bfz_{1:T}^0$ indeed captures only the mutual information, we rely on the clustering task as a proxy.

After the common dynamics are extracted, we utilize Time Series Cluster Kernel (TCK) \citep{tck} to project the multivariate times series onto a lower dimensional space where clustering tasks can be administered.
Specifically, TCK clusters multivariate time series (MTS) by leveraging an ensemble of Gaussian Mixture Models (GMMs) to generate a positive semi-definite kernel matrix, capturing temporal dependencies and enabling kernel-based clustering methods.
If a model is effective enough in extracting the common dynamics, we anticipate a distinct classification boundary.

\begin{figure*}[!t]
\begin{center}
\scalebox{0.92}{
\begin{tabular}{@{}cccccc@{}}
\includegraphics[trim={0cm 0cm 0cm 0cm}, width=.16\linewidth]{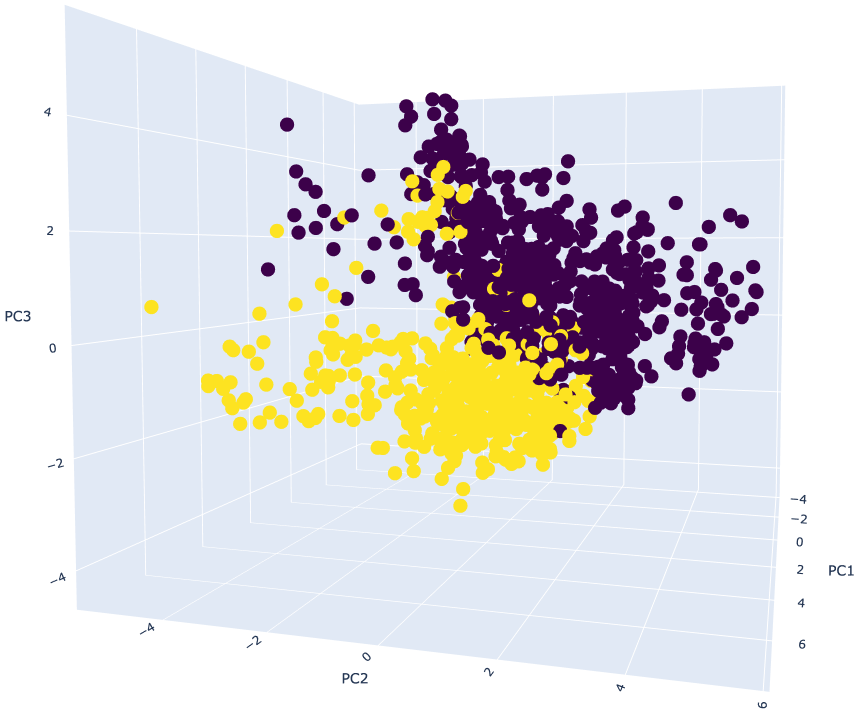}&
\includegraphics[trim={0cm 0cm 0cm 0cm}, width=.16\linewidth]{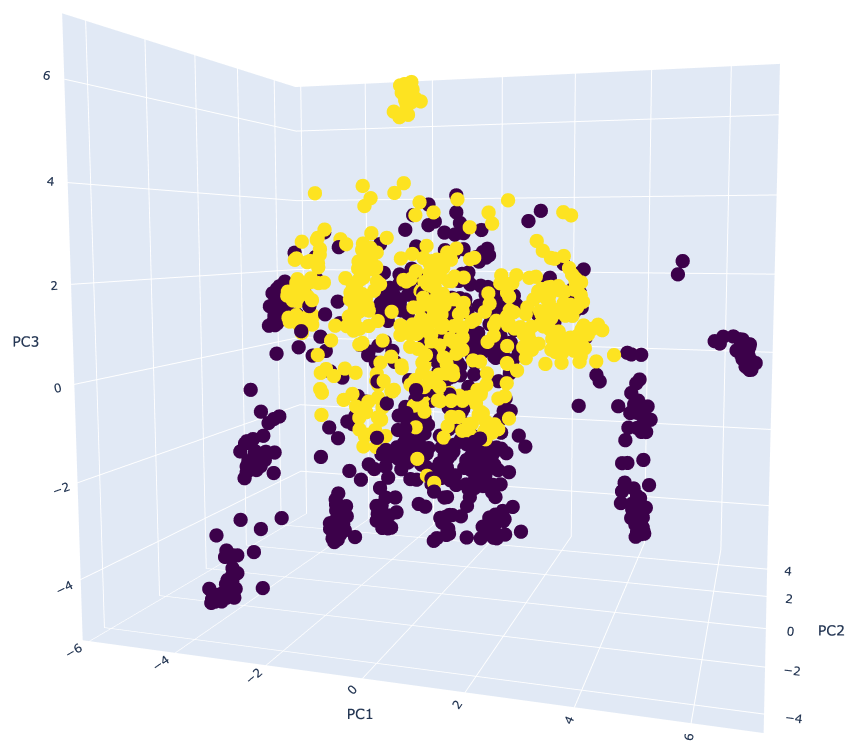}&
\includegraphics[trim={0cm 0cm 0cm 0cm}, width=.16\linewidth]{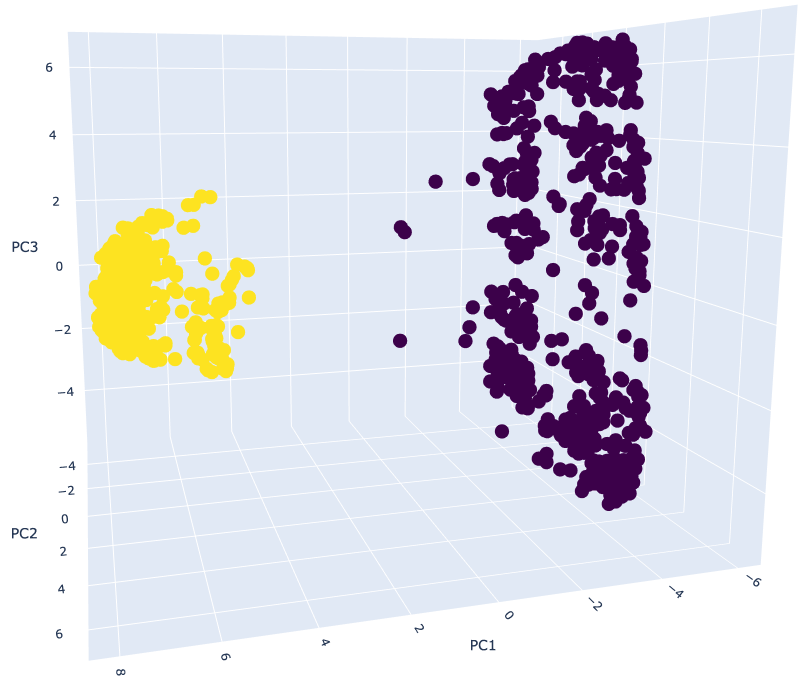}&
\includegraphics[trim={0cm 0cm 0cm 0cm}, width=.16\linewidth]{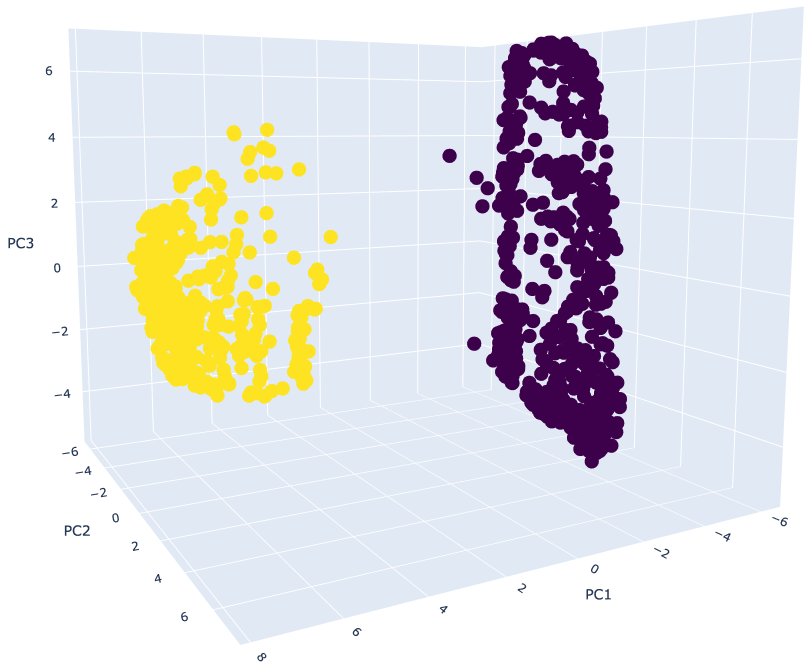}&
\includegraphics[trim={0cm 0cm 0cm 0cm}, width=.16\linewidth]{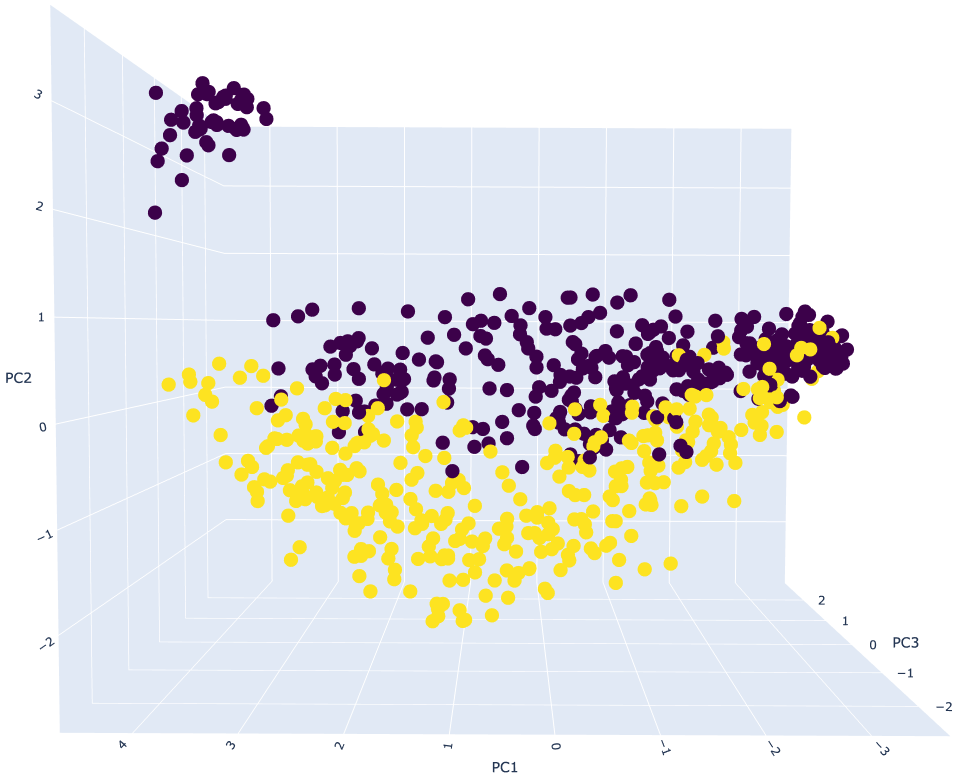}&
\includegraphics[trim={0cm 0cm 0cm 0cm}, width=.16\linewidth]{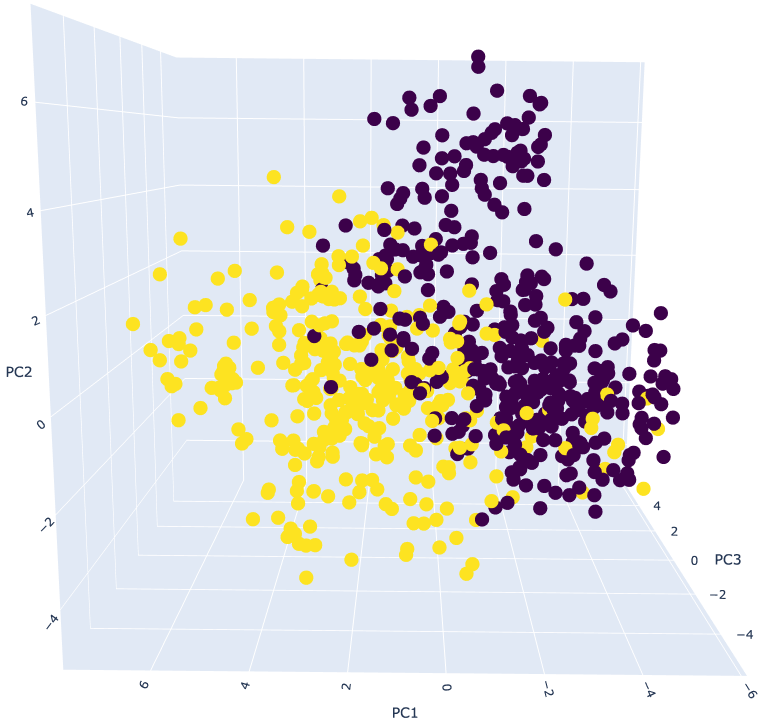}\\
(a) ADNI PCA & (b) ADNI DPCCA  &
(c) ADNI Step I & (d) ADNI Step II  &
(e) NYU PCA & (f) NYU DPCCA \\
\includegraphics[trim={0cm 0cm 0cm 0cm}, width=.16\linewidth]{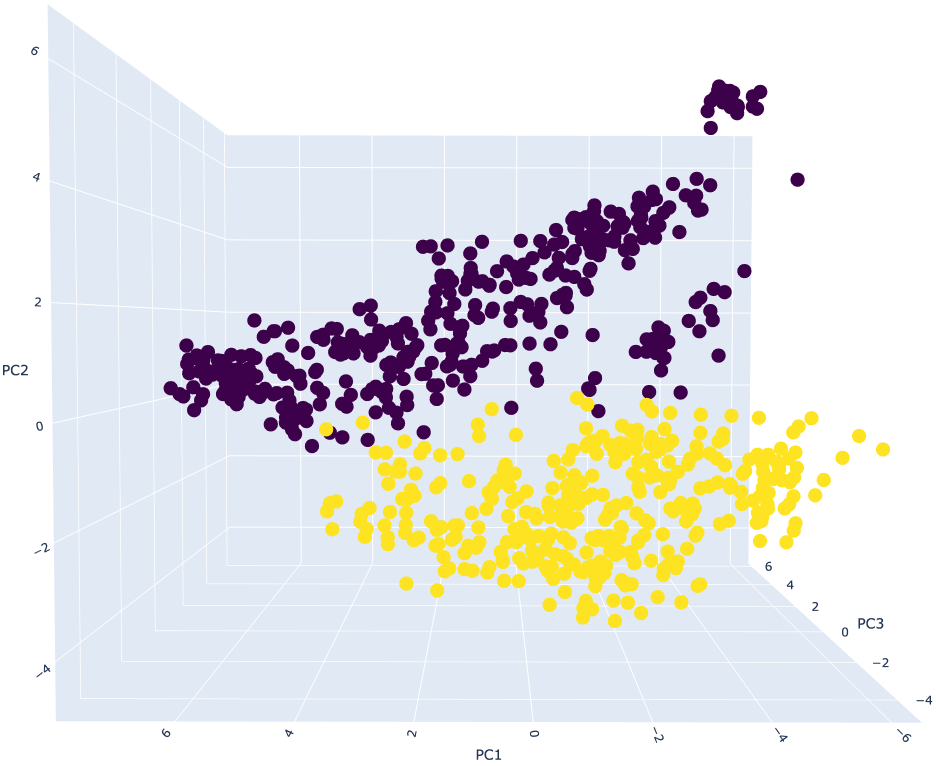}&
\includegraphics[trim={0cm 0cm 0cm 0cm}, width=.16\linewidth]{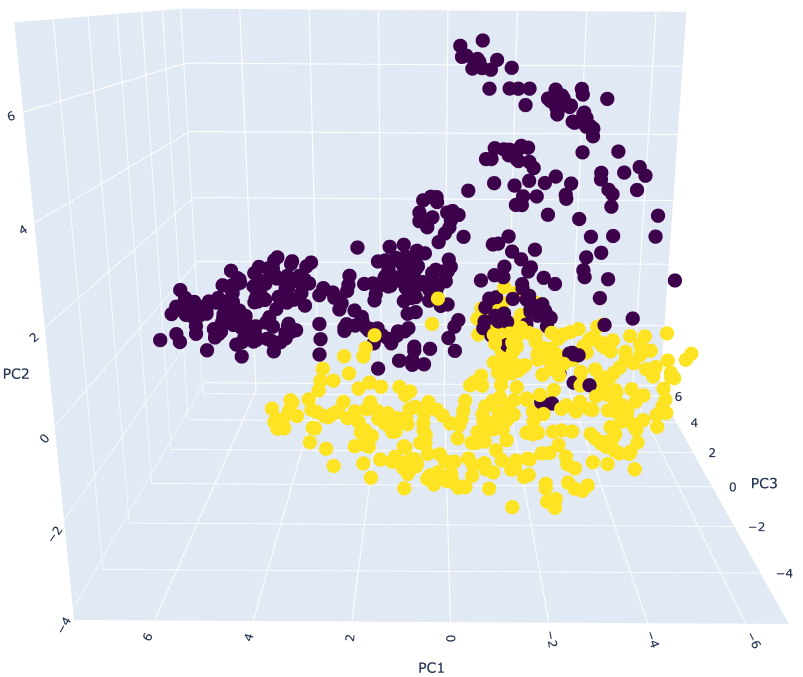}&
\includegraphics[trim={0cm 0cm 0cm 0cm}, width=.16\linewidth]{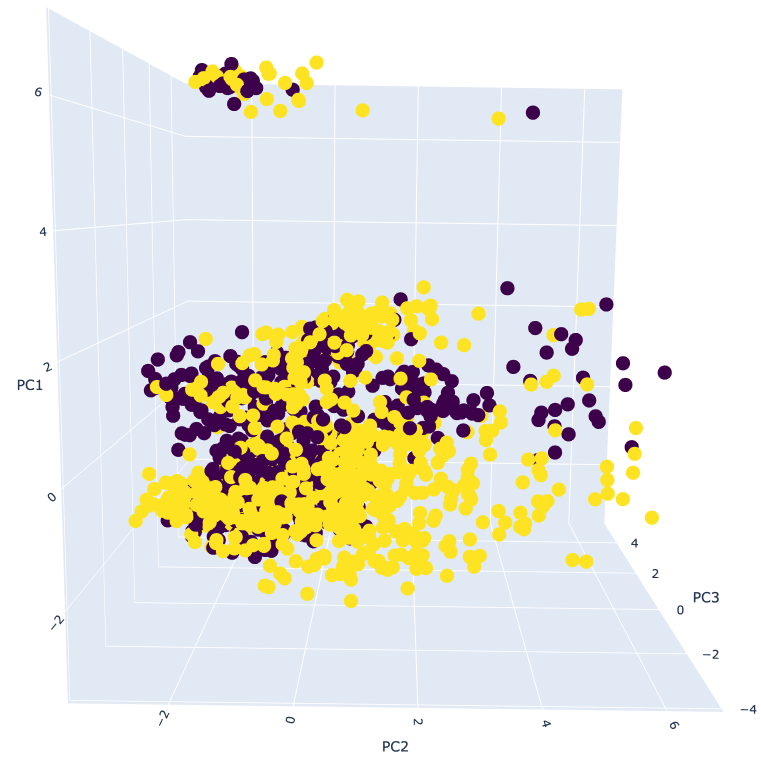}&
\includegraphics[trim={0cm 0cm 0cm 0cm}, width=.16\linewidth]{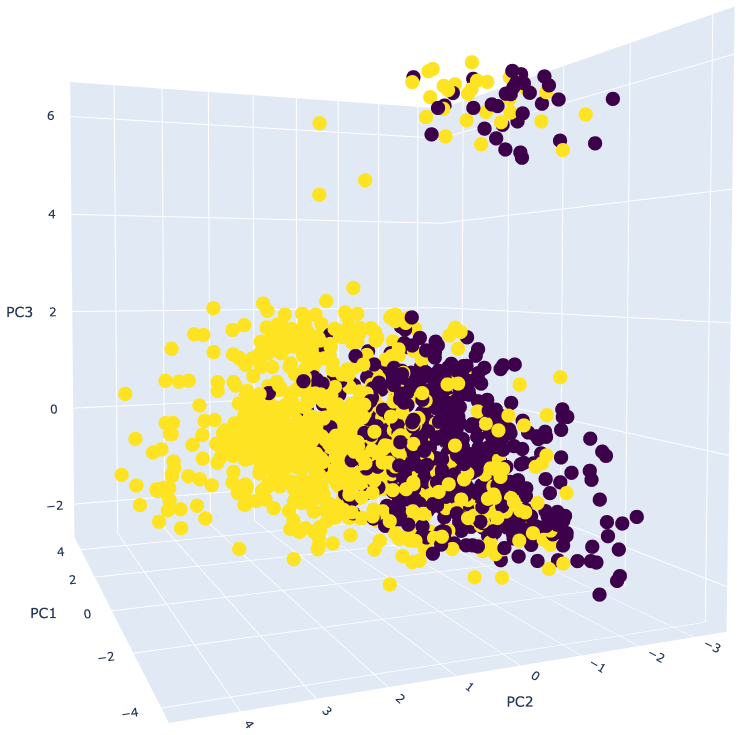}&
\includegraphics[trim={0cm 0cm 0cm 0cm}, width=.16\linewidth]{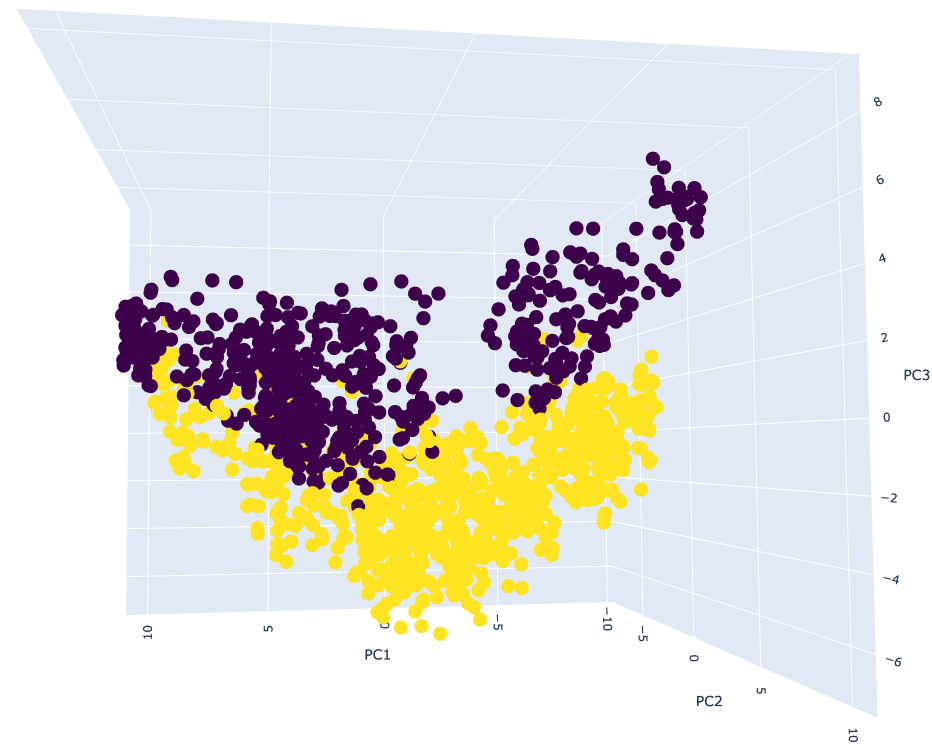}&
\includegraphics[trim={0cm 0cm 0cm 0cm}, width=.16\linewidth]{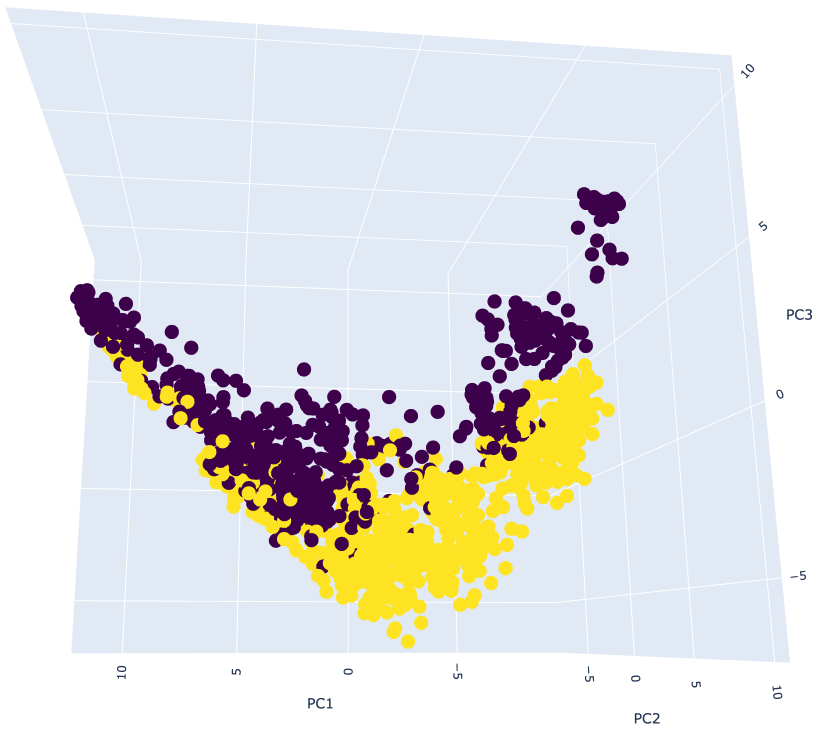}\\
(g) NYU Step I & (h) NYU Step II  &
(i) ECEO PCA & (j) ECEO DPCCA  &
(k) ECEO Step I & (l) ECEO Step II \\
\end{tabular}
}
\end{center}
\setlength{\abovecaptionskip}{0pt}
\caption{Visual inspection of clustering tasks on three datasets (ADNI, NYU, ECEO) using four models (PCA, DPCCA, Step I, and Step II of InfoDPCCA). Subfigures (a)--(d) show results for ADNI, (e)--(h) for NYU, and (i)--(l) for ECEO.}
\label{fig:cluster}
\end{figure*}

Apart from DPCCA and InfoDPCCA, we add two more models for comparison:
\begin{enumerate}
\item PCA: Since directly classifying the
raw data is not an option due to the large dimensions of time series data, we first perform dimension
reductions using PCA before calculating the kernel matrix with TCK.
\item Step I of InfoDPCCA:  The shared latent states sampled from step I can be viewed as their prior distribution $p(\bfz_t^0|\bfx_{1:t}^{1:2})$, while $\bfz_t^0$ sampled after step II can be regarded as posterior distribution $p(\bfz_t^0|\bfx_{1:T}^{1:2})$.
\end{enumerate}


We visualize the extracted latent representations using kernel principal component analysis (KPCA) \citep{kpca} and compare the clustering results of PCA, DPCCA, Step I and II of InfoDPCCA across different datasets. Figure \ref{fig:cluster} presents the 3D scatter plots of the learned representations, where each point corresponds to a subject's extracted latent dynamics, and colors represent different neural states or patient groups.
As shown in the figure, InfoDPCCA models yield a more structured and separable representation space compared to PCA and DPCCA.
This indicates that InfoDPCCA better captures the underlying shared dynamics within each group, leading to improved class separability.
Furthermore, the scatterplots generated by Step I and II of InfoDPCCA are similar, which aligns with the claim that they correspond to the prior and posterior distributions of the latent states.

To perform a quantitative assessment, we apply TCK to the extracted latent states and evaluate the clustering performance using Normalized Mutual Information (NMI) as external metric and Silhouette score as internal metric.
NMI measures the agreement between predicted clusters and true labels, while the Silhouette score evaluates cluster cohesion and separation.
Specifically, we train a classifier on 30\% labeled data and compute clustering metrics on its predictions for the remaining 70\% data.
Table \ref{tab:dpcca_info} summarizes the results, demonstrating that InfoDPCCA models consistently outperform PCA and DPCCA in all three datasets.

\begin{table}
\centering
\begin{tabular}{lccc}
\toprule
 & ADNI & NYU & ECEO \\
\midrule
PCA & 0.592/0.639 & 0.588/0.662 & 0.103/0.514 \\
DPCCA & 0.301/0.587 & 0.417/0.600 & 0.197/0.240 \\
Step I & \textbf{1.000}/0.907 & \textbf{0.865}/\textbf{0.726} & \textbf{0.572}/\textbf{0.662} \\
Step II & \textbf{1.000}/\textbf{0.908} & 0.642/0.620 & 0.228/0.586 \\
\bottomrule
\end{tabular}
\caption{Numeric evaluation of the clustering task on three datasets (ADNI, NYU, ECEO) using four models (PCA, DPCCA, Step I, and Step II of InfoDPCCA). The bolded number indicates the best performer in its respective column. Each cell contains two metrics, NMI/Silhouette score (higher values indicate better performance).}
\label{tab:dpcca_info}
\end{table}

\section{Summary}
\label{sec:summary}

We introduced InfoDPCCA, an information-theoretic exten-

\noindent sion of DPCCA that extracts shared and private latent representations from sequential data. By enforcing an information bottleneck objective and a two-step training scheme, InfoDPCCA improves interpretability and robustness. Experimental results on synthetic and fMRI data demonstrate its effectiveness in capturing meaningful latent structures.

We list a few future directions:
\begin{enumerate}
\item Hyperparameter tuning: Selecting optimal values for $\alpha$ and $\beta$.
\item Evaluation criteria: Developing better metrics beyond correlation coefficients to validate claims.
\item Multiset extension: Extending InfoDPCCA to multiple data streams, similar to DPCCA.
\item Right now, InfoDPCCA can be regarded as an unsupervised approach, which means they are unaware of the labels to each sample in the training process. As opposed to the approach described in \citep{fmri19}, which trains a conditional generative model, InfoDPCCA is not motivated to separate different groups in training. We leave the supervised extension of InfoDPCCA as well as its applications to identifying biomarkers in neuroscience as future work.
\end{enumerate}
These directions will enhance the model's flexibility and applicability.


\begin{acknowledgements} 

This work was funded in part by the Research Council of Norway under grant 309439. The authors would like to express their sincere gratitude to the anonymous reviewers, Dr. Hongzong Li (CityU HK), and Dr. Haoxuan Li (Peking University) for their thorough and constructive feedback, which significantly improved the quality of this work.

\end{acknowledgements}

\bibliography{main_UAI}

\newpage

\onecolumn

\title{Supplementary Material}
\maketitle

\appendix
\section{Related Method}
In this section, we discuss a method related to InfoDPCCA named Multi-View Information Bottleneck (MVIB) \citep{mvib}. Multi-view IB is an extension of the classical Information Bottleneck framework tailored to settings with multiple perspectives of the same underlying entity. Let $\mathbf{v}_1$ and $\mathbf{v}_2$ denote two distinct views of a common object. The primary goal is to derive stochastic mappings $q_\theta(\mathbf{z}_1 | \mathbf{v}_1)$ and $q_\psi(\mathbf{z}_2 | \mathbf{v}_2)$, such that the resulting latent representations $\mathbf{z}_1$ and $\mathbf{z}_2$ encapsulate solely the mutual information $I(\mathbf{v}_1, \mathbf{v}_2)$ shared between the two views. Multi-View IB operates as an unsupervised representation learning technique, with the learned representations being applicable to downstream tasks such as classification.

\begin{figure}[H]
\begin{center}
\begin{tabular}{c}
\includegraphics[clip, width=.5\linewidth]{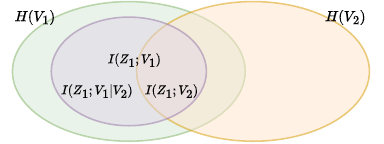}
\end{tabular}
\end{center}
\setlength{\abovecaptionskip}{0pt}
\caption{
Venn diagram illustrating the relationship between the feature of the view $\bfz_1$ and the two views $\bfv_1$ and $\bfv_2$.
}
\label{fig:mvvae}
\end{figure}

Similar to the settings of VIB, Multi-view IB assumes a factorized joint distribution:
\begin{align*}
p(\mathbf{v}_1, \mathbf{v}_2, \mathbf{z}_1, \mathbf{z}_2) = p(\mathbf{v}_1, \mathbf{v}_2) q_\theta(\mathbf{z}_1 | \mathbf{v}_1) q_\psi(\mathbf{z}_2 | \mathbf{v}_2).
\end{align*}
For each view, the optimization objective is defined as:
\begin{align*}
\min \quad I(\mathbf{z}_1; \mathbf{v}_1 | \mathbf{v}_2) - \lambda I(\mathbf{z}_1; \mathbf{v}_2),
\end{align*}
where $I(\cdot; \cdot | \cdot)$ represents conditional mutual information, and $\lambda$ is a trade-off parameter balancing the two terms; an illustration of this objective is shown in Fig. \ref{fig:mvvae}. In contrast to the supervised nature of VIB, which emphasizes label prediction, Multi-View IB seeks to maximize $I(\mathbf{z}_1; \mathbf{v}_2)$. This term captures the shared structure between views, potentially including information not directly relevant to a specific supervised task.

We summarize the similarities and differences between the MVIB and InfoDPCCA, with respective graphical models shown in Fig. \ref{fig:venn_z0} and Fig. \ref{fig:mvvae}:
\begin{itemize}
\item The features learned from MVIB, $\bfz_1$ and $\bfz_2$, are representations of observations $\bfv_1$ and $\bfv_2$ respectively; while in InfoDPCCA, the common latent factor $\bfz^0$ is a representation of both observations $\bfx^{1:2}$.
\item The two methods have the same goal of learning representations that capture  the mutual information between the observations while discarding the redundant information.
\item InfoDPCCA uses mainly variational approximations, while MVIB relies on auxiliary mutual information estimators.
\end{itemize}

\section{Experimental Setup on Numerical Simulation}
The synthetic dataset is generated using the HÃ©non map, a chaotic dynamical system, to model a $2$-dimensional ground truth latent $\bfz_t \in \R^2$. For each of $1000$ sequences, the latent state evolves over $T=300$ time steps via the HÃ©non map:
$$\bfz_{t+1} = [1-1.4 z_{t,0}^2 + z_{t,1}, 0.3 z_{t,0}]^\top,$$
with initial condition $z_{0,0} \in \mathrm{Unif}(-1,1)$ and $z_{0,1} \in \mathrm{Unif}(-0.1,0.1)$. High-dimensional observations $\bfx_t, \bfy_t \in \R^{120}$ are then created by projecting $\bfz_t$ through random linear transformation $W_x, W_y \in \R^{120 \times 2}$ (drawn from $\Normal(0,1)$), followed by additive Gaussian noise:
$$\bfx_t = W_x \bfz_t + \epsilon_x,\quad \bfy_t = W_y \bfz_t + \epsilon_y,$$
where $\epsilon_x, \epsilon_y \in \Normal(0,0.05^2 I_{120})$. The dataset, comprising $1000$ sequences of paired $(\bfx_t, \bfy_t)$ with corresponding $\bfz_t$, is split into 80\% training and 20\% testing subsets. The global-mean correlation metric in Equation \eqref{eq:corr} measures a model's ability to capture shared latent dynamics by averaging correlation coefficient $\rho_{jk}$ between true latent states $z_{ntj}$ ($j$-th dimension of $\bfz_n$ at $t$-th time step) and estimated states $\hat{z}_{ntk}$ across all sequences $n\in [N]$ and time steps $t\in [T]$.

\section{Additional Experiments on fMRI Dataset}
\begin{figure}[ht]
\begin{center}
\begin{tabular}{@{}cc@{}}
\includegraphics[trim={0cm 0cm 0cm 0cm}, width=.3\linewidth]{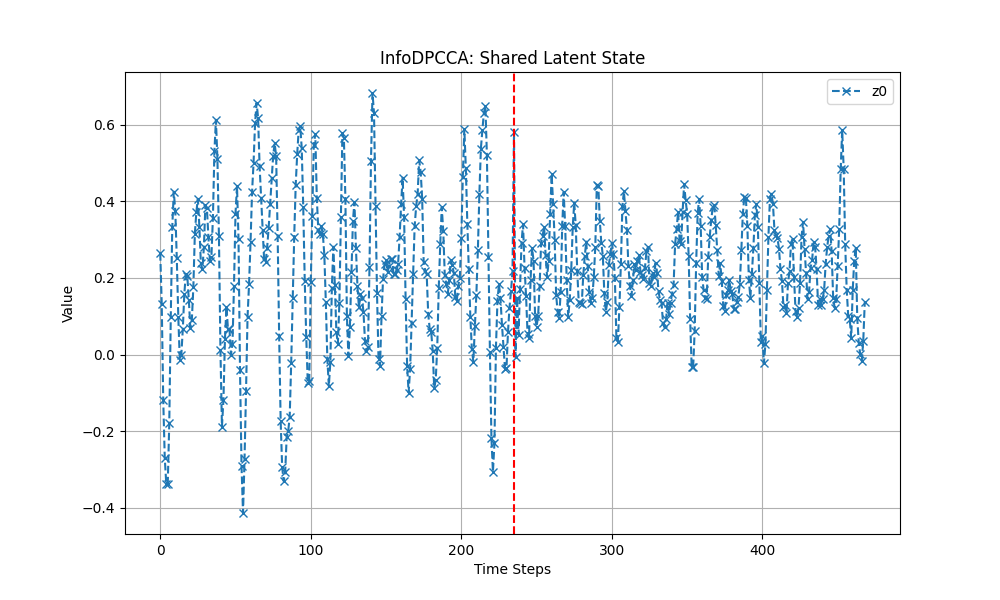} &\includegraphics[trim={0cm 0cm 0cm 0cm}, width=.3\linewidth]{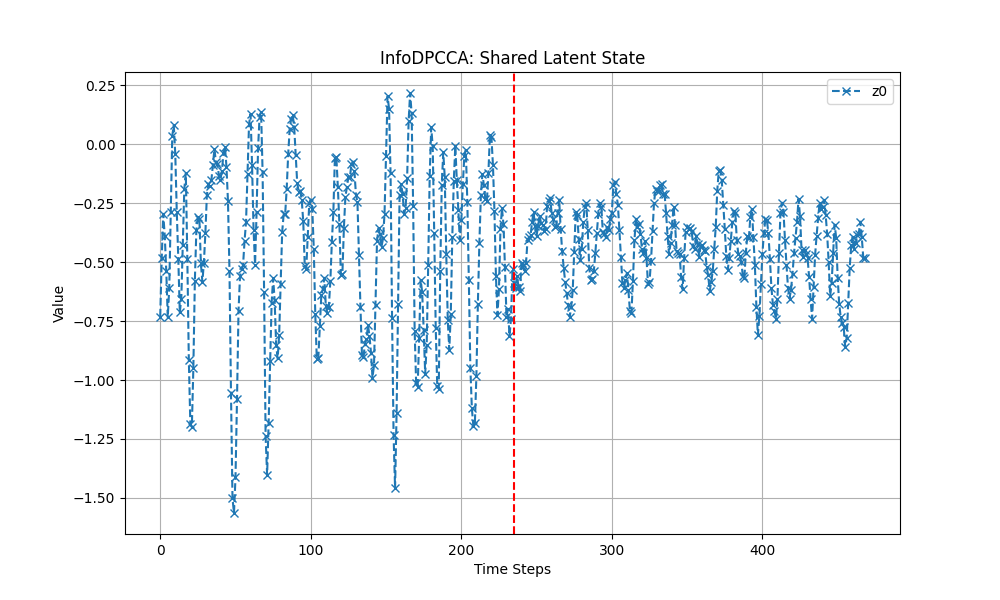}\\
(a) & (b) \\
\includegraphics[trim={0cm 0cm 0cm 0cm}, width=.3\linewidth]{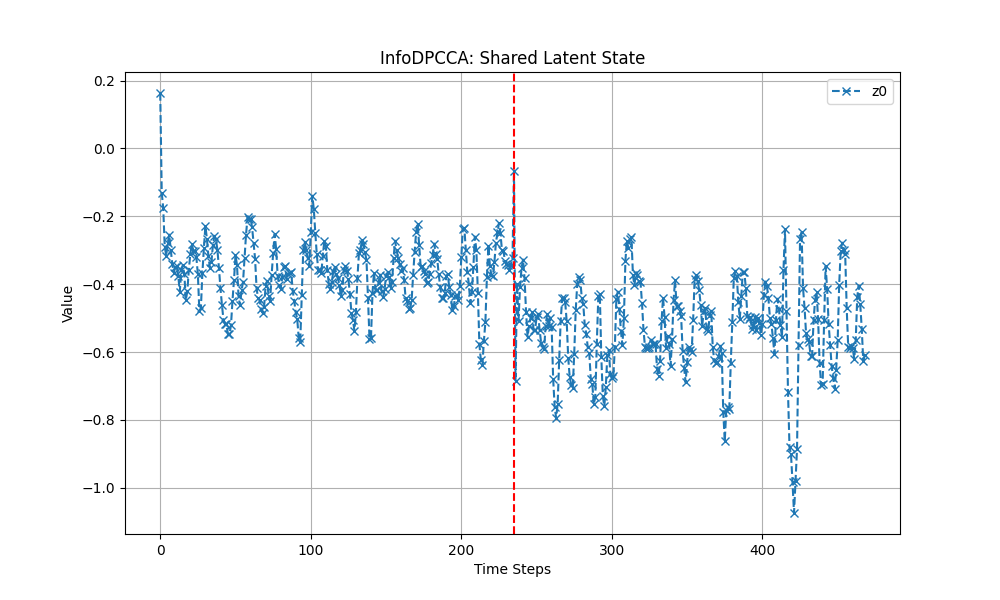} &\includegraphics[trim={0cm 0cm 0cm 0cm}, width=.3\linewidth]{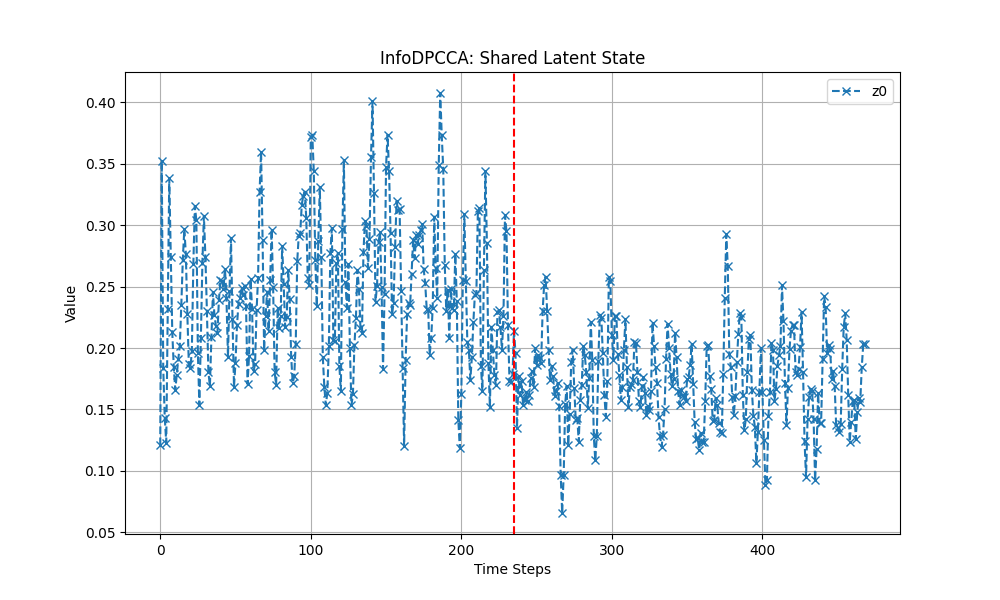}\\
(c) & (d)
\end{tabular}
\end{center}
\setlength{\abovecaptionskip}{0pt}
\caption{Extracted shared latent states for the supplementary experiment on the ECEO dataset.}
\label{fig:sup_exp}
\end{figure}

As mentioned in the paper, the ECEO dataset contains fMRI data of 48 subjects recorded with their eyes open and closed. We extract the shared latent states between EC and EO of the same subject, and we hypothesize that if the extracted shared latent states are identity-specific, they should be able to detect the change in identity. To make our setting more clear, we use math notations. Suppose we have dataset $\cD = \{\EC_i, \EO_i: i \in [48]\}$. We denote $x_i^1 = \concat(\EC_i, \EC_{2i+1})$ and $x_i^2 = \concat(\EO_i, \EO_{2i+1})$. We then transform the dataset to $\cD' = \{x_i^1, x_i^2: i\in [24]\}$ and use it to train InfoDPCCA. We found that the shared latent states of InfoDPCCA are able to detect the change of identity by:
\begin{itemize}
\item Altering the magnitudes: Figure \ref{fig:sup_exp} (a) and (b).
\item Shifting the means: Figure \ref{fig:sup_exp} (c) and (d).
\end{itemize}

\end{document}